\documentclass[10pt,journal,compsoc]{IEEEtran}
\usepackage[table,dvipsnames]{xcolor}
\usepackage[table]{xcolor}
\RequirePackage[dvipsnames]{xcolor}
\usepackage{amsmath,amsfonts}
\usepackage{algorithmic}
\usepackage[ruled,vlined]{algorithm2e}

\usepackage{array}
\usepackage[caption=false,font=normalsize,labelfont=sf,textfont=sf]{subfig}
\ifCLASSOPTIONcaptionsoff
 \usepackage[nomarkers]{endfloat}
\let\MYoriglatexcaption\caption
\renewcommand{\caption}[2][\relax]{\MYoriglatexcaption[#2]{#2}}
\fi
\usepackage{textcomp}
\usepackage{stfloats}
\usepackage{url}
\usepackage{verbatim}
\usepackage{graphicx}
\def\BibTeX{{\rm B\kern-.05em{\sc i\kern-.025em b}\kern-.08em
    T\kern-.1667em\lower.7ex\hbox{E}\kern-.125emX}}
\usepackage{balance}
\usepackage[caption=false,font=normalsize,labelfont=sf,textfont=sf]{subfig}
\usepackage{textcomp}
\usepackage{stfloats}
\usepackage{url}
\usepackage{verbatim}
\usepackage{graphicx}
\usepackage{soul}
\usepackage{tcolorbox}
\usepackage{algorithmic}

\usepackage{algorithm}

\usepackage{longtable}
\usepackage{supertabular}
\ifCLASSOPTIONcompsoc
  \usepackage[nocompress]{cite}
\else
  \usepackage{cite}
\fi
\usepackage{color}
\usepackage[normalem]{ulem}  
\usepackage{xspace}  
\usepackage{booktabs}
\usepackage{multirow}
\usepackage{pifont}
\usepackage{arydshln}
\usepackage{colortbl}
\usepackage{hyperref}
\usepackage{makecell,rotating,multirow,diagbox}
\makeatletter
\DeclareRobustCommand\onedot{\futurelet\@let@token\@onedot}
\def\@onedot{\ifx\@let@token.\else.\null\fi\xspace}
\makeatother
\def\eg{\emph{e.g.}} 
\def\ie{\emph{i.e}\onedot}

\newcommand{\extmf}[1]{\textcolor{black}{#1}}
\newcommand{\hhh}[1]{\textcolor{black}{#1}}
\renewcommand{\st}[1]{\textcolor{black}{#1}}

\begin{document}

\title{Self-Consistency as a Free Lunch: Reducing Hallucinations in Vision-Language Models via Self-Reflection}


\author{Mingfei~Han*\thanks{*Authors contribute equally.},
        Haihong~Hao*,
        Jinxing Zhou,
        Zhihui~Li, 
        Yuhui~Zheng,
        Xueqing~Deng,
        Linjie~Yang,
        Xiaojun~Chang,~\IEEEmembership{Senior~Member,~IEEE}
\IEEEcompsocitemizethanks{
\IEEEcompsocthanksitem Mingfei Han and Jinxing Zhou are with the Department of Computer Vision, Mohamed Bin Zayed University of Artificial Intelligence, Abu Dhabi, United Arab Emirates. E-mail: \{hmf282@gmail.com, zhoujxhfut@gmail.com\}.
\IEEEcompsocthanksitem Haihong Hao, Zhihui Li and Xiaojun Chang are with the School of Information Science and Technology, University of Science and Technology of China. E-mail: \{haohaihong0529@gmail.com, lizhihuics@ustc.edu.cn, xjchang@ustc.edu.cn\}
\IEEEcompsocthanksitem Xueqing Deng and Linjie Yang are with Bytedance Seed, San Jose, USA. \protect \\
E-mail: \{xueqingdeng, linjie.yang\}@bytedance.com.
\IEEEcompsocthanksitem Yuhui Zheng is with School of Computer Science, Qinghai Normal University.
\IEEEcompsocthanksitem The work was done during Mingfei Han's internship at Bytedance.

}}

\markboth{IEEE Transactions on Pattern Analysis and Machine Intelligence}%
{Han \MakeLowercase{\textit{et al.}}: Self-Consistency as a Free Lunch: Reducing Hallucinations in Vision-Language Models via Self-Reflection}


\IEEEtitleabstractindextext{%
    \begin{abstract}
Vision-language models often hallucinate details, generating non-existent objects or inaccurate attributes that compromise output reliability. 
Existing methods typically address these issues via extensive human annotations or external supervision from more powerful models.
In this work, we present a novel framework that leverages the model’s self-consistency between long responses and short answers, to generate preference pairs for training. 
We observe that short binary questions tend to yield highly reliable responses, which can be used to query the target model to evaluate and rank its generated responses.
\extmf{Specifically, we design a self-reflection pipeline where detailed model responses are compared against concise binary answers, and inconsistency signals are utilized to automatically curate high-quality training data without human annotations or external model-based supervision. }
By relying solely on self-consistency rather than external supervision, our method offers a scalable and efficient solution that effectively reduces hallucinations using unlabeled data. 
Extensive experiments on multiple benchmarks, \ie, AMBER, MultiObject-Hal (ROPE), Object HalBench and MMHal-Bench, demonstrate significant improvements in factual grounding and reliability. 
Moreover, our approach maintains robust instruction-following ability, as evidenced by enhanced performance on LLaVA-Bench and MMBench.

\end{abstract} 
}


\maketitle

\section{Introduction} \label{sec:intro} 
\IEEEPARstart{V}{ision}-Language Models (VLMs)~\cite{antol2015vqa} have achieved remarkable performance across tasks such as image captioning, visual question answering, and scene understanding, largely owing to extensive pretraining on massive text corpora that endow them with rich world knowledge. 
\extmf{Models such as LLaVA~\cite{llava15}, Qwen2.5-VL~\cite{bai2025qwen2}, and GPT-4V~\cite{GPT4V} have exhibited unprecedented abilities in interpreting intricate visual semantics and responding to complex multi-turn instructions. Nonetheless, despite these capabilities, 
current VLMs frequently generate \emph{hallucinations}, the details in their outputs that are not grounded in the visual inputs~\cite{rohrbach2018object,wang2024amber,biten2021let,li2023evaluating}. Such inaccuracies pose serious challenges in safety-critical scenarios, for instance, medical diagnostics, autonomous driving, and high-stakes surveillance systems.}


\extmf{Hallucinations in VLMs manifest across multiple semantic layers, including incorrectly identifying objects absent from images (object hallucinations), assigning incorrect properties to visible objects (attribute hallucinations), and falsely describing spatial or semantic relationships (relational hallucinations)~\cite{wang2024amber,rohrbach2018object}. 
At a fundamental level, these hallucinations inherently stem from the large-scale vision-language pretraining process itself.
Typical vision-language pretraining datasets inevitably contain varying degrees of noise, subjective image descriptions, or human labeling errors, which are hard to fully eliminate given the scale and diversity of web-collected data~\cite{shao2024visualcot,marino2019ok}. 
Additionally, pretrained language models bring intrinsic linguistic biases and priors formed through exposure to extensive textual corpora where certain linguistic expressions commonly co-occur. 
These biases predispose models to prioritize fluent narrative structures and popular linguistic patterns over accurate visual grounding. 
For instance, even when an image does not clearly depict a common contextual association (``bananas are yellow'', ``cats sit on couches''), models may hallucinate such plausible but incorrect visual details. 
Thus, eliminating hallucinations remains an open and challenging problem, fundamentally linked to dataset imperfections, linguistic priors, and inherent labeling ambiguities. These inherent complexities highlight that developing practical, scalable, and effective hallucination mitigation methods continues to be an important and actively studied research direction.
}

\begin{figure}[t!] 
    \centering
    \includegraphics[width=\linewidth]{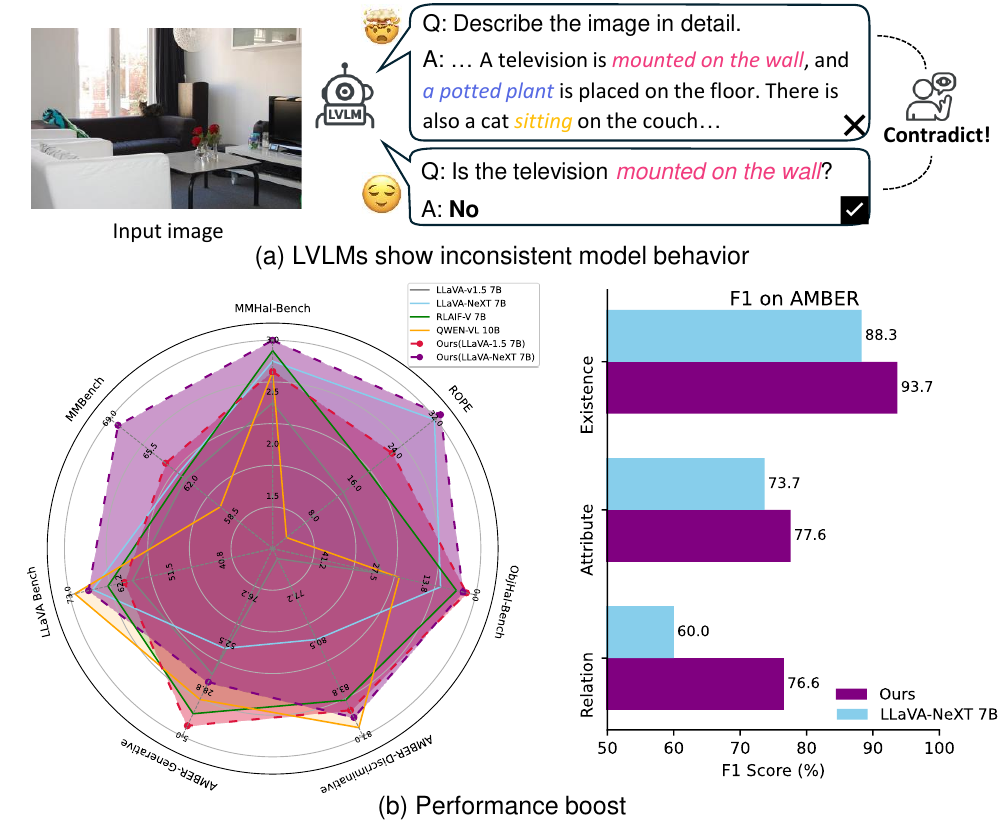}
    \vspace{-7mm}

    \caption{\textbf{Motivating observation and overall performance.} 
    (a) Illustration of hallucination versus self-consistency. The base model (LLaVA-v1.5) hallucinates in the detailed caption, where each semantic aspects are highlighted with colors; whereas, the model correctly denies the hallucination when explicitly queried with short binary questions. This inconsistency motivates our idea of exploiting self-consistency signals.
    (b) Performance radar comparison. Our method substantially outperforms the baselines (\eg, LLaVA‑v1.5) and prior methods across hallucination benchmarks; \eg, on AMBER, discriminative F1 improves across all categories, with the overall score rising from \textbf{74.7\% to 85.2\%}, while the hallucination rate drops \textbf{from 36.4\% to 10.3\%}. It also improves helpfulness on general benchmarks, \ie, MMBench and LLaVA-Bench, achieved without external training supervision.} 
    \vspace{-4mm}

    \label{fig:motivation}
\end{figure}

Prior work has attempted to mitigate hallucinations through various approaches. Some methods rely primarily on improving decoding strategies, such as contrastive decoding~\cite{VCD,manevich2024mitigating,kim2024code}, logit calibration~\cite{favero2024multi}, or generative feedback~\cite{zhang2025self} to directly suppress hallucinated tokens \extmf{at generation time. 
These methods are intuitive, however, fail to directly correct the underlying semantic ambiguities that drive hallucinations.}
Others employ reinforcement learning (RL)-based techniques, such as RLAIF~\cite{yu2024rlaif} and MOCHa~\cite{ben2023mitigating}, or iterative revision mechanisms~\cite{zhou2024calibrated} in which the model refines its outputs through self-feedback or multi-agent debates. External critics~\cite{zhou2023analyzing,yin2024woodpecker}, such as object detectors~\cite{yin2024woodpecker}, have also been used to verify generated content. 
However, these strategies often rely on ``blind'' scoring procedures~\cite{liu2023mitigating} that are inconsistent and difficult to control~\cite{yan2024evaluating}, or they depend on external signals that may not well align with human judgments, both of which introduce additional complexity and computational overhead.

Instead of retrieving supervision signals from external models, we are motivated to use self-reflection as hallucination indicator.
\extmf{Specifically, VLMs seem prone to hallucinate when generating elaborate visual elements, yet tend to produce significantly more accurate responses to simple atomic questions. 
As illustrated in Figure~\ref{fig:motivation}(a), when the VLM (\eg, LLaVA) generates a detailed caption describing the scene, it might incorrectly state ``a television is mounted on the wall'' and ``a potted plant is placed on the floor'' despite the factual correctness in the visual scene. 
Surprisingly, the same model, when explicitly queried with short yes/no questions about the identical semantic aspect, \ie, ``Is the television mounted on the wall?'', provides contradictory yet visually accurate binary answers (``No'').
This intriguing inconsistency highlights a crucial observation: short binary visual queries, due to minimal linguistic complexity, tend to produce fewer hallucinations. We hypothesize that the internal consistency between short answers and detailed descriptions, agreement or contradiction, can serve as a scalable, annotation-free training signal inherently different from expensive external feedback used previously.
Motivated by this insight, we propose exploiting such self-consistency to reliably and effectively mitigate hallucinations in VLMs.
}

In this paper, we introduce \textbf{Self-reflection Preference Optimization}, a novel framework designed to enhance the reliability of VLMs by leveraging self-consistency. Our approach comprises three key phases: \textbf{Hallucination exposure}: The model generates detailed descriptions, and we then query atomic facts in these outputs using concise, yes/no questions to expose potential hallucinations. 
\textbf{Self-reflection preference ranking}: We measure the consistency between the detailed outputs and the corresponding short responses, rating them based on the frequency of inconsistent atomic facts. 
\textbf{Self-reflection preference optimization}: Using the generated ratings, the model is fine-tuned under a Direct Preference Optimization framework to favor more factually consistent responses to improve consistency and factual grounding.
Our approach derives a self-supervised signal directly from the model’s own responses, thereby avoiding the pitfalls of ``blind'' scoring and reducing computational overhead.
\extmf{Although self-reflected methods for generating preference data, such as GRPO~\cite{shao2024deepseekmath}, have been recently popular in language reasoning tasks, they fundamentally differ from our multimodal consistency-based approach which does not depend on external reward verifiers.}

\hhh{To evaluate our framework, we conduct experiments across diverse and challenging benchmarks.
Specifically, }\extmf{as illustrated in Figure~\ref{fig:motivation}(b),
equipped with Self-reflection Preference Optimization, our models clearly and consistently outperform previous established hallucination-mitigation methods across multiple challenging benchmarks, including Object HalBench~\cite{li2023evaluating}, MMHal-Bench~\cite{sun2023aligning}, and Multi-Object ROPE~\cite{chen2025multi}. Additionally, on AMBER~\cite{wang2024amber}, our mitigation approach substantially improves hallucination measurements related to object existence, attributes, and relations, elevating overall discriminative F1 scores from 74.7\% to 85.2\% with a marked decrease in hallucination rates (from 36.4\% down to 10.3\%), compared to the baseline LLaVA-v1.5~\cite{liu2023llava}.} 
\extmf{Moreover, we validate the superiority and generalizability of our approach on more advanced vision-language models, including LLaVA-NeXT-7B~\cite{llavanext}, LLaVA-LLaMA3-8B~\cite{llavanext}, and Qwen2.5-VL-7B~\cite{bai2025qwen2}. Beyond hallucination mitigation, our approach also demonstrates notable improvements in model helpfulness across widely recognized benchmarks such as MMBench~\cite{liu2024mmbench} and LLaVA-Bench~\cite{liu2023llava}, without relying on external training supervision. These results collectively show the effectiveness, robustness, and broad applicability of our self-reflection preference optimization method.}

To summarize, our contributions are as follows: 
\vspace{-1mm}
\begin{itemize} 
    \item \extmf{We propose a novel self-reflective mechanism harnessing VLMs’ inherent consistency between detailed response and atomic short-answer queries. Crucially, this internal consistency signal serves as scalable, annotation-free supervision, fundamentally different from existing methods relying on human annotations or external feedback models.}
    \item \extmf{We introduce Self-reflection Preference Optimization, a preference-based optimization framework explicitly designed to leverage internal self-consistency signals for robustly mitigating diverse hallucination types in vision-language tasks.}
    \item \extmf{Extensive experiments demonstrate that our method significantly surpasses existing state-of-the-art approaches across multiple benchmarks (\eg, Object HalBench, MMHal-Bench, Multi-Object ROPE). Particularly for AMBER discriminative tasks, our method notably improves discriminative F1 (74.7\%$\rightarrow$85.2\%) while sharply reducing hallucination rates (36.4\%$\rightarrow$10.3\%).}
\end{itemize}

\section{Related Works}
\label{sec:related}



\subsection{\extmf{Vision-Language Hallucinations}}
Hallucination is a common issue in Vision-Language Models (VLMs)
, where the models tend to generate responses that align with patterns from their training data rather than relying on actual visual input. This can lead to misinterpretations of the visual context, diminishing trustworthiness and accuracy in applications ranging from image captioning to visual question answering. Studies~\cite{wang2024amber,rohrbach2018object,biten2021let,liu2023mitigating} have shown that hallucination in VLMs often manifests as the inclusion of incorrect objects, attributes, or scene descriptions, which are not present in the visual input. 
Researchers are actively working to alleviate hallucinations. For example,
VCD~\cite{VCD} and ICD~\cite{wang2024mitigating} utilize distortions of the original visual input to expose hallucinations and guide models to focus on visual evidence. 
Ha-DPO~\cite{ha_dpo} and Povid~\cite{zhou2024aligning} generate negative samples containing hallucinations, enabling models to actively recognize and avoid errors. 
Less-is-more~\cite{yue2024less} reduces the hallucination rate by restricting responses to highly confident objects.
LLaVA-RLHF~\cite{sun2023aligning} trains the model on human-annotated preference data to mitigate hallucinations.
In addition to alleviating hallucinations, existing research also focuses on comprehensively and objectively evaluating hallucination from multiple perspectives, as well as categorizing hallucinations into different types.
AMBER~\cite{wang2024amber} evaluates hallucinations in both generative and discriminative tasks. ROPE~\cite{chen2025multi} focuses on multi-object hallucinations. Despite these efforts, VLMs still face significant challenges regarding hallucination generation.

\subsection{External feedback for hallucination detection}
To address uncertainty and over-reliance on biased patterns in training data, several works focus on detecting hallucinations with external feedback within its outputs. First type of feedback is through Reinforcement Learning from Human Feedback (RLHF) \cite{sun2023aligning} to guide the models respond based on visual-grounded cues. Silkie \cite{2023vlfeedback} utilizes the powerful GPT-4V~\cite{GPT4V} to generate feedback to help it generate visually faithful responses. A recent method RLAIF-V~\cite{yu2024rlaif} incorporate a divide-and-conquer strategy, where responses are split into atomic claims, allowing models to evaluate the validity of each component. This approach assesses the validity of each claim using either an external powerful VLM or the original model with binary questions. Then the responses are rated using the produced validity scores and used to generate training pairs for Direct Policy Optimization (DPO)~\cite{rafailov2023dpo}. 
\hhh{This approach relies on peer feedback—using other open-source MLLMs as the feedback source. While it avoids dependence on powerful external superior models used in superior-teacher methods, it still requires additional models and cannot reduce hallucinations on its own.
}
In contrast to existing work, we focus on harvesting feedback from the target model itself, which does not require expansive human annotation or powerful external models. 

\subsection{VLM reasoning and Chain-of-Thought} 
\label{sec:related_cot}
Chain-of-Thought (CoT)~\cite{wei2023chainofthought} is a powerful prompting technique that enables LLMs to conduct reasoning before the final response. Researchers also utilize CoT to improve VLMs on complex tasks such as graphical math problems~\cite{zhang2024mavis} and localization tasks~\cite{shao2024visualcot}. A recent work~\cite{zhang2024visioncot} uses ground truth answers to rate model-generated reasoning chains, to further improve the VLM's reasoning capability using reinforcement learning. We employed a similar CoT method to generate detailed responses for QA tasks, and evaluate the validity of the response with our self-reflective verification to rate the quality of each response, without ground truth annotation for the visual questions.

\hhh{
\subsection{Self-Feedback Models}
Like humans reflect on and analyze their prior answers to improve them, large language models can also self-feedback using their own previous outputs without additional supervised training data\cite{wang2024eaco,wang2024enhancing,madaan2023self,zelikman2022star,yuan2024self}. SIMA\cite{wang2024enhancing} generates multiple diverse responses and then evaluates all self-generated responses using carefully designed critic prompts, classifying them as Positive Response or Negative Response. SELF-REFINE\cite{madaan2023self} generates an initial output with an LLM, then the same LLM critiques that output and uses the feedback to iteratively refine itself.
Self-Taught Reasoner \cite{zelikman2022star} and Self-Rewarding\cite{yuan2024self} use chain-of-thought prompts to guide an LLM to expose potential hallucinations in its reasoning process, and then let the model itself evaluate and select the better response. All of these methods employ feedback from the target model itself and secure notable performance gains. However, hallucination detection is far harder than simple right-or-wrong judgments in multiple-choice, and self-assessment may hallucinate repeatedly. Our method defines concrete steps for exposing hallucinations, instead of relying on the model’s implicit self-assessment. Moreover, existing approaches offer no clear metric for comparing output reliability, whereas we use inconsistent model behaviors as an explicit criterion for evaluation.
}

\section{VLM Hallucinations and Behavioral Consistency}
\label{sec:behavior_consistency}

\extmf{In this section, we first characterize hallucinations across multiple dimensions and introduce the phenomenon of behavioral consistency, where discrepancies between a model's open-ended responses and short, binary-question answers can reliably indicate hallucinated content. We further quantitatively validate this correlation through a careful manual check, establishing a precise, scalable signal for identifying hallucinations and thus motivating our subsequent self-reflection framework as in Section~\ref{sec:methods}.}


\subsection{\textcolor{black}{Hallucinations in VLMs}}
\begin{figure*}[t!] 
    \centering
    \includegraphics[width=\linewidth]{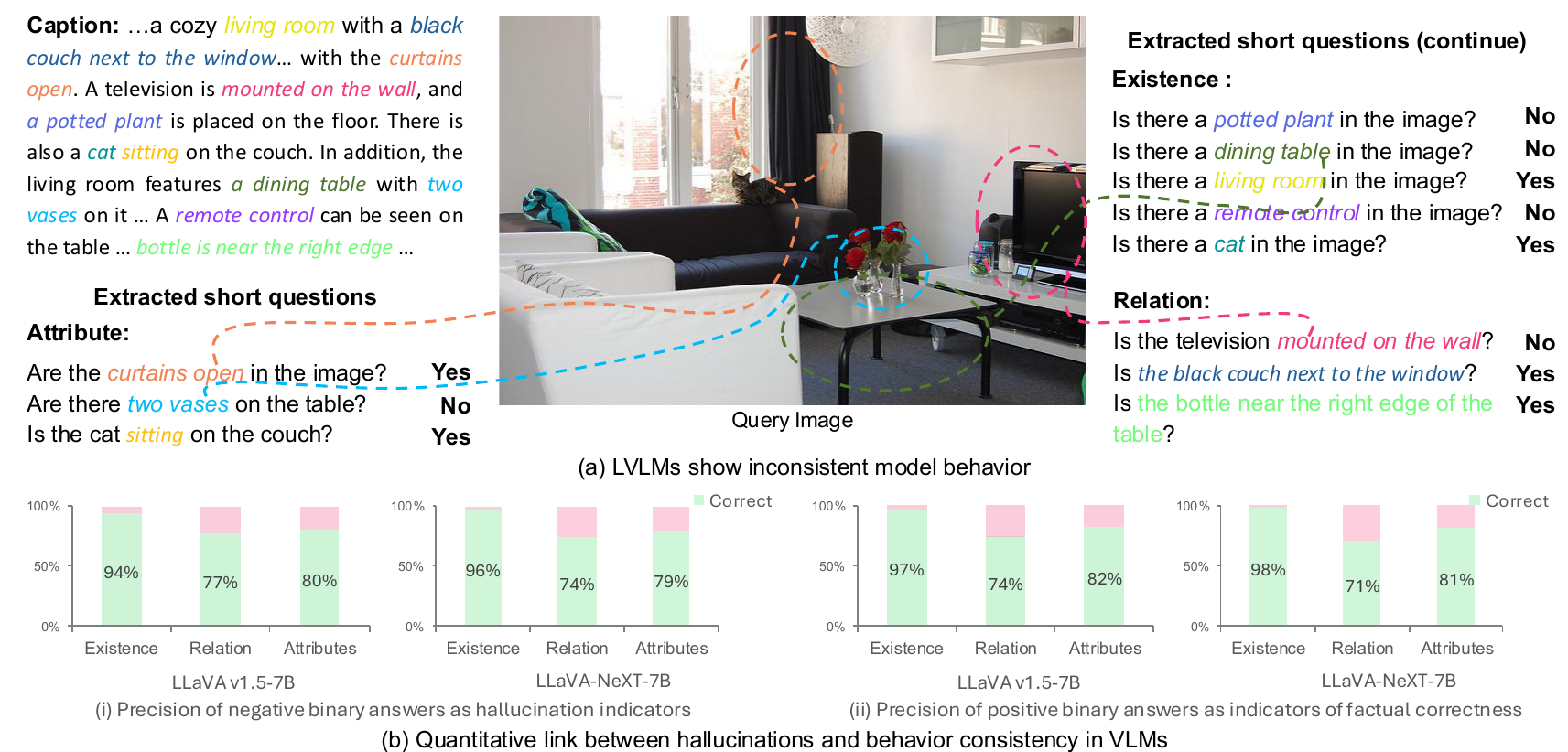}

    \caption{\textbf{Observation of inconsistent model behaviors.} 
    \textbf{(a)} Caption generated by LVLMs often hallucinates, for example, containing a description of \textcolor{Periwinkle}{\textit{potted plant}} which doesn't exist. However, when querying with a short question, \ie, \textit{Is there a potted plant in the image?}, the same model may respond with ``No'', denying the existence. We show different short-questions extracted from image captions, categorized into \textit{Existence}, \textit{Relation} and \textit{Attributes}. Different colors in questions correspond to relevant facts in the image caption. Some inconsistency are also highlighted on the input image with dotted circles.
    \textbf{(b)} We manually check 100 short questions per model (LLaVA-v1.5-7B~\cite{llava15}, LLaVA-NeXT-7B~\cite{llavanext}). We report the precision of negative answers and positive answers in two panels separately. \extmf{Green bars mark correct judgements. The consistently high precision across categories and models confirms that short-questioning is a lightweight yet reliable signal for identifying hallucinated content.}
    Detailed analysis could be found in Section~\ref{sec:behavior_consistency}.
    }

    \label{fig:observation}
\end{figure*}

Vision-Language Models (VLMs) frequently generate hallucinations, \ie, the outputs containing content not grounded in visual input. Following recent benchmarks like AMBER~\cite{wang2024amber} and CHAIR~\cite{rohrbach2018object}, we categorize the hallucinations into three distinct types. (1) Object-level hallucinations: introducing objects that are visually absent.
(2) Attribute-level hallucinations: assigning incorrect attributes (\eg, actions, quantity, size, shape, color, or material) to visible objects.
(3) Relational hallucinations: incorrectly describing spatial or semantic relationships, such as non-existent physical contact between objects or false positional relationships.
Figure~\ref{fig:observation} visually illustrates various different occasions of these hallucination types using concrete examples from common VLMs (\eg, LLaVA-v1.5 and LLaVA-NeXT), highlighting the subtlety and variety of such errors. Understanding this detailed taxonomy provides a necessary foundation for investigating the link between hallucination and behavioral consistency, as we explore in subsequent sections.

\subsection{Behavior Consistency in VLMs}

\extmf{Despite the prevalence of hallucinations described above, an intriguing phenomenon emerges when probing VLMs with queries of different types: the behavior consistency gap.
Specifically, behavior consistency, as explored in this work, refers to the agreement between detailed responses (\ie, lengthy captions or comprehensive answers) and short, binary questions posed to the same model. 
Intuitively, a model-generated detailed description may contain hallucinated details, whereas short binary responses to targeted questions tend to be more accurate. }

\extmf{As shown in Figure~\ref{fig:observation}(a), 
the VLM-generated caption inaccurately describes ``a potted plant is placed on the floor'', ``the living room features a dining table'' thereby introducing a hallucinated ``dining table'' and ``potted plant on the floor''. These hallucinations might be due to lack of spatial perception (\eg, tables surrounded by couches in a living room typically are not dining tables), or confusion caused by object occlusion (\eg, the black stool near the television stand, partially occluded by the coffee table and vases). 
However, when subsequently queried, ``Is there a dining table in the image?'' or ``Is there a potted plant in the image?'', the model responds confidently both with ``No'', denying its previously hallucinated content.
Such contradiction indicates an inconsistency in VLM behaviors: detailed captions may hallucinate freely, while the same model often reveals accurate visual grounding when it is restricted to binary short judgements. Intuitively, the gap arises because binary questions are inherently easier for VLMs to answer correctly, with short binary answers directly and explicitly grounding the visual content in the image. Conversely, long open-ended image captions convey more freedom for VLMs to hallucinate, driven by the learned bias and linguistic priors from training data.}

\extmf{Inspired by this discrepancy, we propose to leverage this naturally emerged behavior consistency gap as a scalable, cost-free indicator to identify hallucinated elements in VLM outputs, without external supervision. In the following subsection, we quantitatively validate this correlation through careful manual checks.}



\subsection{Quantitative Linking}

\extmf{To quantitatively validate the correlation between the observed behavior consistency phenomenon and hallucinations, we conducted a careful manual evaluation.
For clarity, we first assume the atomic facts are accurately extracted from the detailed model-generated responses, which is elaborated in~\ref{sec:data_curation}.
These atomic facts are then transformed into short binary questions, categorized three predefined types, \ie, \textit{Object Existence} (querying whether the object of interest is present in the image), \textit{Attributes} (verifying attributes of objects, such as color, quantities, and actions) and \textit{Relations} (checking spatial or direct contact relationships among objects). 
Specifically, a binary answer of ``No'' indicates that the queried fact is absent from the image and thus hallucinated in the detailed response; while a ``Yes'' indicates the queried fact is visually grounded and accurately described.}

\extmf{We randomly sampled 100 queries from outputs of VLMs (\eg, LLaVA-v1.5-7B~\cite{llava15} and LLaVA-NeXT-7B~\cite{llavanext}) and independently assessed whether these binary responses correctly aligned with visual grounding. 
To ensure comprehensive validation, we separately evaluated precision for negative and positive binary answers.
As shown in Figure~\ref{fig:observation}(b)(i), negative responses (answered with \emph{No}) from LLaVA-v1.5-7B demonstrated remarkably high precision as hallucination indicators, 94\% for Object-Existence, 77\% for Relations, and 80\% for Attributes elements. 
This highlights the reliability of using the negative response as an indicator of hallucinations.
Conversely, 
positive binary responses strongly correlated with correctly grounded visual elements, with precision scores of 97\%, 74\%, and 82\% for Object-Existence, Relations, and Attributes respectively.
On the more advanced LLaVA-NeXT-7B~\cite{llavanext}, we observed similar effectiveness, reinforcing the consistency of this correlation across models.}

\extmf{These results confirm our empirical hypothesis: the robust correlation between internal consistency (as measured by short binary responses) and factual correctness supports its utility as an effective indicator of response quality, enabling us to reliably rank model-generated responses based on their factual grounding. 
Based on this finding, 
model-generated responses can subsequently be ranked by their consistency scores. Section~\ref{sec:data_curation} details how the consistency indicator is computed and effectively utilized in ranking.}

\begin{figure*}[ht!] 
    \centering
    \includegraphics[width=\textwidth]{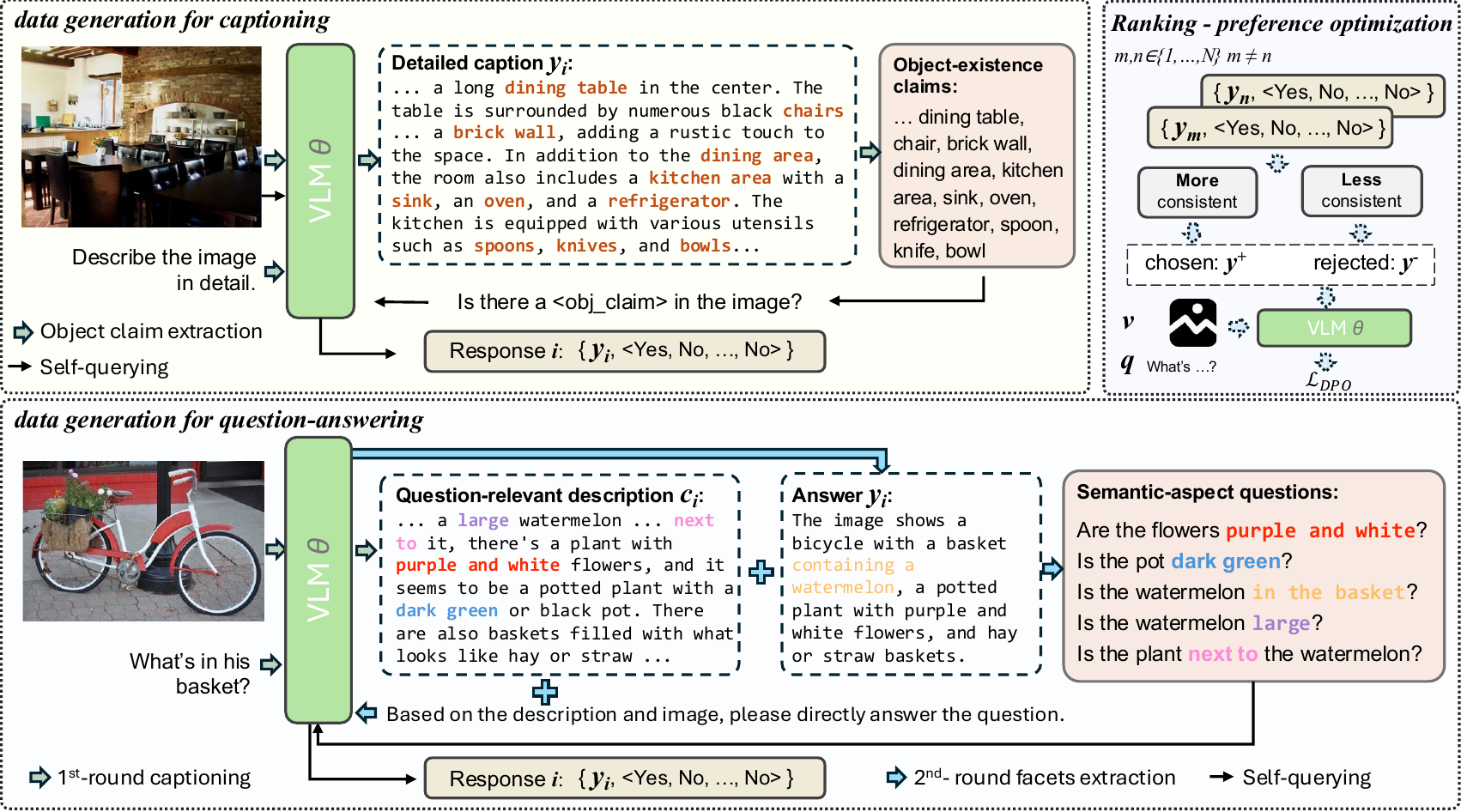}
    \caption{
    Pipeline of our self-consistent preference optimization, consisting of data generation (captioning and question-answering) and ranking for preference optimization. Colored arrows indicate specific steps (see legend). 
    \textbf{Top-left:} the captioning data generation, generating detailed captions and extracting object-existence claims which are subsequently converted into short binary questions for VLM self-querying consistency.
    \textbf{Bottom:} the question-answering data generation. A question-relevant caption is first generated and subsequently used in a two-round CoT prompting approach to produce the final answer. 
    Then both the caption and the final answer are analyzed to extract atomic claims related to semantic aspects, forming short questions used for consistency checking.
    \textbf{Top-right:} ranking procedure used to construct preference data pairs. For each input \{$v$,$q$\}, $N$ candidate responses are generated; 
    for any two sampled responses \{$y_m$,$y_n$\}, the more consistent response (fewer ``No'' responses) is chosen as preferred, while the less consistent (more ``No'' responses) is rejected.
    }
    \label{fig:pipeline}
\end{figure*}

\section{Self-reflection Preference Optimization}
\label{sec:methods}


\extmf{Motivated by our findings in Section~\ref{sec:behavior_consistency}, we propose to leverage the behavioral consistency phenomenon as a scalable and self-contained supervision signal to mitigate hallucinations in VLMs. Specifically, we first introduce a self-reflection data-curation pipeline (Section~\ref{sec:data_curation} and Algorithm~\ref{alg:alg1}) that automatically generates binary short questions from atomic facts extracted from VLM-generated detailed responses. We then utilize the inherent consistency between these binary answers and detailed responses to generate high-quality preference pairs. Finally, we incorporate these preference pairs into a Direct Preference Optimization (DPO)~\cite{rafailov2023dpo}-based fine-tuning objective (Section~\ref{sec:method_optimize}), effectively reducing hallucination without additional human labeling or external critics.}

\subsection{Self-reflection Data Curation}
\label{sec:data_curation}

\extmf{To enable scalable and effective self-reflection supervision, we curate preference training data directly from model-generated responses. We first run a pre-trained VLM (\eg, LLaVA-v1.5-7B~\cite{llava15}) on a set of unlabeled images to produce detailed responses, including captioning and open-ended question-answering. For each of them, we automatically extract fine-grained atomic facts, defined as minimal assertions (\eg, an object's existence, attributes, or relation to others).}

\renewcommand{\algorithmicrequire}{\textbf{Input:}}
\renewcommand{\algorithmicensure}{\textbf{Output:}}
\begin{algorithm}[t!]
\caption{Self-reflection data curation (single pass). We repeat $N$ times with preference ranking to construct the training data for preference optimization.}\label{alg:alg1}
\begin{algorithmic}
\REQUIRE Image set $\mathcal{V}=\{v_i\}_{i=1}^n$, where $\mathcal{V} =$ captioning data $ \mathcal{V}_{\mathrm{cap}} \cup $ question-answering data $\mathcal{V}_{\mathrm{qa}}$, pretrained VLM $\theta$
    \IF{$v_i \in \mathcal{V}_{\mathrm{cap}}$}
        \STATE $x_i \leftarrow$ “Describe the image in detail.”
    \ELSIF{$v_i \in \mathcal{V}_{\mathrm{qa}}$ }
        \STATE $x_i \leftarrow$ corresponding question
    \ENDIF
\ENSURE Self-reflection Data $I= \{I_i\}_{i=1}^n$. \\$I_i=(y_i, \{a_{i,1}, ..., a_{i,m_i}\}),a_{i,j} \in \{\mathrm{yes},\, \mathrm{no}\},j\in\{1,..., m_i\}$\\
\STATE {\textcolor{gray}{ /* Begin process */}}
\FOR{$i = 1$ to $n$}
    \IF{$v_i \in \mathcal{V}_{\mathrm{cap}}$}
        \STATE Caption $y_i \leftarrow \text{ModelResponse}(\theta, v_i, x_i)$\\
        \STATE Claims $\{o_{i,j}\} \leftarrow \text{ExistenceClaimExtractor}(y_i)$\\
        \STATE Binary questions $\{bq_{i,j}\} \leftarrow \{\text{QuestTemplate}(o_{i,j})\}$\\
        \STATE \{$a_{i,j}\} \leftarrow \{\text{ModelResponse}(\theta, v_i, bq_{i,j})\}$\\
    \ELSIF{$v_i \in \mathcal{V}_{\mathrm{qa}}$}
        \STATE {\textcolor{gray}{/* 1st step of QA response */}}
        \STATE $x_{c} \leftarrow$ ``Describe the image.''
        \STATE Description $c_i \leftarrow \text{ModelResponse}(\theta, v_i, x_{c})$\\
        \STATE {\textcolor{gray}{/* 2nd step of QA response */}}
        \STATE Answer $y_i \leftarrow \text{ModelResponse}(\theta, v_i, [x_i;c_i])$\\
        \STATE {\textcolor{gray}{/* Semantic claims and questions extraction */}}
        \STATE $\{bq_{i,j}\} \leftarrow \text{QuestionGenerator2}(y_i)$\\
        \STATE \{$a_{i,j}\} \leftarrow \{\text{ModelResponse}(\theta, v_i, bq_{i,j})\}$
    \ENDIF\\
    $I_i=(y_i, \{a_{i,j}\}_{j=1}^{m_i})$
\ENDFOR
\RETURN $I$
\end{algorithmic}
\label{alg:self-reflection}
\end{algorithm}

\noindent\textbf{Captioning \textcolor{black}{Dataset}.} 
Given an image $v$ and a textual prompt $x$ (\eg \textit{``Describe this image in detail''}), a VLM (parametrized as $\theta$) generates multiple detailed captions ${y_{1}, y_{2}, \dots, y_{N}}$ via $y \sim p_{\theta}(\cdot \mid x, v)$.
For each caption $y_{i}$, we extract \textit{object-existence claims}, such as \textit{knife}.
We automatically convert each claim into binary queries using a predefined template (\eg, \textit{``Is there a knife in the image?''}).
We then re-query the same VLM with these short binary questions, evaluating internal consistency by the occurrence of negative (``No'') answers, which indicate inconsistent model behavior.
Although we initially explored extracting claims related to semantic aspects 
(\eg, objects' attributes, relations), experimental validation showed that object-existence claims provided the most effective signal.

\noindent\textbf{Question-answering \textcolor{black}{Dataset}.} In the VQA task, the response from the VLM is usually short. It is almost impossible to directly extract enough atomic facts from the short responses to be evaluated by the self-consistency approach. In order to gather more detailed responses with abundant factual information, we employ a two-round CoT prompting approach. Given an image $v$ and a question $q$, we first prompt the model to generate a description $c$ that is relevant to the question.
The model subsequently generates a final answer $y$ conditioned on both the image and the preliminary description:
$y \sim p_{\theta}(\cdot \mid q, c, v)$.
From the model responses \{$c$,$y$\},
we then extract atomic claims and binary short questions that correlated to \st{semantic aspects} 
(\eg, attributes, relations). 
The same VLM is then prompted again with these short questions to verify the correctness of each facet. Similar to the captioning task, we measure internal consistency based on the presence of negative responses.
Empirical studies demonstrate that this two-round process significantly outperforms a single-round approach, highlighting that the preliminary description notably clarifies context and reduces ambiguity.

\extmf{For object-existence claims, we follow the same practice from AMBER~\cite{wang2024amber}} to extract substantive nouns as atomic claims.
\extmf{For semantic claims correlated with object relations and attributes, we employ GPT-4 with a structured prompt (\textcolor{black}{see supplementary material}) to return a JSON object response, which we then parse to obtain atomic facts. 
The prompt specifies the task, the required output format, and several in‑context examples.
Each JSON entry contains the original source text, the claim category, and a properly formatted binary question.
To ensure every extracted claim preserves the wording and scope of the input, we explicitly state this constraint in the prompt, demonstrate it with in-context examples, and require the original text to be returned in the output. As a result, no additional judgment and verification is introduced in this process.}

\noindent\textbf{Preference Data Generation and Ranking.}
In practice, we generate three candidate responses (\ie, $N$=3) for each question or prompt. 
To effectively mitigate hallucinations, we construct preference pairs by measuring factual inconsistency of each candidate response, explicitly counting the number of inconsistent claims identified by negative binary answers to their corresponding short questions. Intuitively, a greater number of \emph{No} responses indicates a greater occurrence of hallucinated content, and therefore implies a less preferable candidate response.
\extmf{We explore two ranking strategies based on the number of inconsistencies, \ie, \textit{Occurrence Ranking}: rank candidates directly by the number of inconsistent claims $K$; fewer inconsistencies implies higher preference; \textit{Relative Ratio Ranking}: rank by the ratio of inconsistencies to total claims, \ie, $K/T$, where $T$ is the number of atomic facts in the response.}

\extmf{Although both ranking strategies are reasonable, we adopt \textit{Occurrence Ranking} as find strategy for generating preference data, as it yields better performance in downstream hallucination mitigation, as discussed in Section~\ref{sec:ablations}.}

\noindent\textbf{Discussion:} A closely-related method to our approach is RLAIF-V~\cite{yu2024rlaif}, which also extracts sub-questions from responses, but relies on more advanced VLMs (\eg, LLaVA-NeXT-34B) for factual verification. In contrast, our approach relies on behavioral consistency within the same VLM, forming a self-contained pipeline without external scoring and ensuring greater transparency and scalability. 
Moreover, RLAIF-V utilizes Best-of-N (BoN) = 32 at inference, whereas we use single-shot greedy decoding (1$\times$ runtime) for fairness. BoN sampling is orthogonal to both methods.





\subsection{Self-reflection Preference Optimization}
\label{sec:method_optimize}
With our pre-curated preference pairs obtained from internal consistency signals, we then optimize the VLM parameters using DPO~\cite{rafailov2023dpo}.
Specifically, for input pair $(x,v)$, we have preference pairs comprising a preferred response $y^+$ and a less-preferred response $y^-$. 
The DPO objective is formulated as:
\begin{equation}
\begin{aligned}
\mathcal{L}_{\text{DPO}}(\theta) = -\mathbb{E}_{(x,v,y^+,y^-)} &\biggl[\log \sigma\biggl(\beta\biggl(\log \frac{p_\theta(y^+|x,v)}{p_{\text{ref}}(y^+|x,v)} \\
&- \log \frac{p_{\theta}(y^-|x,v)}{p_{\text{ref}}(y^-|x,v)}\biggr)\biggr)\biggr],
\end{aligned}
\end{equation}
where $p_{\text{ref}}$ denotes the reference (pretrained) VLM $\theta$, $\beta$ is a hyperparameter scaling factor, and $\sigma$ is the sigmoid function. This optimization process aligns the model's output distribution towards internally consistent and preferred outputs.

\noindent\textbf{Training Procedure and Prompts.}
The preference dataset used in training is prepared using the consistency-based ranking method described in Section~\ref{sec:data_curation}, combining data from both captioning and QA tasks. For captioning, we use the ShareGPT4V~\cite{chen2024sharegpt4v} captioning prompt; while for QA, we utilize the original question prompt. Note that for QA data, only the final response $y$ is selected as the target response during VLM fine-tuning. 
Subsequently, our DPO training directly aligns the model with these internally less inconsistent preferences, \ie, less hallucination, without requiring iterative refinement or additional external supervision.

\section{Experiments}
\label{sec:experiments}


\subsection{Implementation Details}
\noindent\textbf{Training data}. Our training data for captioning task is sourced from ShareGPT-4V's COCO\cite{lin2014microsoft} subset, comprising a total of 50k images. Note that only the images are used without their collected captions. Question-answering data is collected from a combination of VQA-v2~\cite{goyal2017making}, OKVQA~\cite{marino2019ok}, TextVQA~
\cite{singh2019towards}, and GQA~\cite{hudson2019gqa}, resulting in 18,328 image-question pairs. Again, no ground truth answers are used in our method.
After self-reflection data curation, we obtain 26,720 and 19,844 preference data pairs for captioning and QA data on LLaVA-v1.5-7B, and 52,182 and 43,161 pairs on LLaVA-NeXT-7B. 
The preference data is sourced from the same image and question collections but differs across models due to variations in generated responses and inconsistency patterns. In addition, we discard pairs where both responses exhibit the same number of inconsistencies to ensure meaningful preference ranking.

\noindent\textbf{Models and implementation.} We fine-tune \textbf{LLaVA-v1.5-7B}~\cite{llava15} using official LLaVA repository~\cite{llava15} and trl library~\cite{vonwerra2022trl} using LoRA~\cite{hu2022lora} (rank=32, dropout=0.05) for parameter-efficient fine-tuning. The training is conducted with an accumulated batch size of 8$\times$4$\times$1 (\#GPUs$\times$\#data-per-GPU$\times$\#gradient-accumulation-steps) and a learning rate of 2e-6, following a cosine decay schedule with a warm-up ratio of 0.03 for 2 epochs.
 We fine-tune \textbf{LLaAV-NeXT-7B}~\cite{llavanext}, \textbf{LLaAV-LLaMA3}~\cite{llavanext} and \textbf{Qwen2.5-VL-7B~\cite{bai2025qwen2}} using LLaMAFactory\cite{zheng2024llamafactory} with LoRA~\cite{hu2022lora}. 
 \extmf{For LLaAV-NeXT-7B, we use rank=32, alpha=64, an accumulated batch size of 8$\times$2$\times$4, a learning rate of 5e-6, a cosine decay schedule with a warm-up ratio of 0.1 for 2 epochs.
For LLaVA-LLaMA3-8B, we use the same LoRA setting (rank=32, alpha=64) but a larger accumulated batch size 8$\times$2$\times$8 and a smaller learning rate of 1e-6, with the same cosine schedule and warm‑up for 2 epochs. 
For Qwen2.5‑VL‑7B, we use rank=8, alpha=16, an accumulated batch size 8$\times$2$\times$4, a learning rate of 5e-6, the same cosine schedule with a warm-up ratio of 0.1 for 3 epochs.}



\subsection{Evaluation and Metrics} 
\extmf{For a thorough analysis,} we evaluate two aspects of the model's performance: trustworthiness and helpfulness, following practice from RLAIF~\cite{yu2024rlaif}. 
To reflect the effectiveness in alleviating vision-language hallucination, 
we adopt ROPE \cite{chen2025multi}, Object HalBench \cite{rohrbach2018object}, AMBER \cite{wang2024amber} and MMHal-Bench~\cite{sun2023aligning} as our trustworthiness benchmarks. 
To reflect the ability in general communication,
we adopt LLaVA-Bench \cite{liu2023llava} and MMBench \cite{liu2024mmbench} as helpfulness benchmarks.

\textbf{Object HalBench} \cite{rohrbach2018object} is a widely used benchmark for evaluating object hallucination in image description. \textcolor{black}{It evaluates captions generated on MSCOCO to quantify both response-level and mention-level hallucination rates across 300 image instances. }We follow the previous works \cite{pi2025strengthening, yu2024rlaif} to apply 8 diverse prompts to obtain detailed and stable image captions (generative task). We report the response-level (percentage of hallucinated sentences, denoted as \textbf{Resp.}) and mention-level hallucination rate (the percentage of hallucinated objects, denoted as \textbf{Ment.}). Both metrics are the lower, the better.

\textbf{ROPE} \cite{chen2025multi} focuses on multiple objects hallucination. VLMs suffer more hallucinations when focusing on multiple objects \cite{mckee2021multi,li2023evaluating}. We use the Default Multi-Object setting 
and follow \cite{liu2024enhancing} to report averaged accuracy on \textbf{Wild} (mixed object classes), \textbf{Hom} (same object class) and \textbf{Het} (tested objects are of different classes) on the unseen split. All metrics are higher, the better.

\textbf{MMHal-Bench} \cite{llavarlhf} consists of 96 challenging questions based on images sourced from OpenImages, each accompanied by ground-truth answers and image contents. Generated responses from the VLM for image-question pairs are fed to GPT-4\cite{GPT4V} to obtain scores, which reflect accuracy and hallucination rates. The GPT-4 scoring ranges from 0 to 6, where a score less than or equal to 2 indicates the presence of hallucinations. The \textbf{Score} represents the average score, with higher values indicating better responses, while \textbf{Hal} denotes the proportion of responses containing hallucinations; a lower value indicates higher response accuracy.

\textbf{AMBER} \cite{wang2024amber} 
measures hallucinations in both generative task (captions task) and discriminative task (QA task). 
The discriminative task measures hallucinations in existence, relation and attributes (including number, state and action). 
We report the accuracy (\textbf{Acc.}) and \textbf{F1} metric in overall and additionally precision and recall metrics for separate types. Both of them are the higher the better.
In the generative task, we report \textbf{CHAIR}~\cite{rohrbach2018object}, \textbf{Cover} (object coverage), \textbf{Hal} (proportion of responses with hallucinations) and \textbf{Cog} (the similarity on hallucinatory target objects between VLM and human) following \cite{wang2024amber}. For CHAIR and Hal, lower is better. For Cog and Cover, higher is better.

\textbf{LLaVA-Bench} \cite{liu2023llava} is used to assess the models' performance in real-world multimodal conversation, detailed description and complex reasoning. It also reflects the instruction-following capability of the model. We use the llava-bench-in-the-wild section, which contains 60 questions in total, to report the overall score of conversation description and reasoning, following the existing setting~\cite{pi2025strengthening}.

\textbf{MMBench}~\cite{liu2024mmbench} 
offers a comprehensive evaluation framework that rigorously tests a model's capabilities across a wide range of complex tasks, including object recognition, reasoning, and scene understanding. 
It adopts circular evaluation process where the model outputs are iteratively refined to ensure consistency and accuracy. 
We report the overall performance on the MMBench-EN-Dev split.



\begin{table*}[htp!]
\caption{Main results. We compare results with methods based on different types of feedback. 
We evaluate object hallucination at different granularities, including response-level (Rsp., Hal.) and mention-level (Men.). 
For multi-object hallucination, we report object category accuracies across different dataset conditions: Wild (mixed object categories), Hom (same object category), and Het (completely different object categories).
Notably, our method using LLaVA-NeXT-7B outperforms all other 7B models, even those leveraging feedback from more powerful models or human annotations.
Additionally, our model based on LLaVA-v1.5-7B achieves the lowest object hallucination rates while maintaining competitive performance across all other benchmarks.}
\vspace{-2mm}
\label{tab:main}
    \centering
    \small
    \resizebox{\textwidth}{!}{
\setlength{\tabcolsep}{0.8mm}
\begin{tabular}{c|cc|cc|ccc|cc|cccc|cc}
\toprule
 \multirow{2}{*}{\textbf{Model}}  & \multirow{2}{*}{\textbf{LLM Size}} & \multirow{2}{*}{\textbf{Feedback }}& \multicolumn{2}{c|}{\textbf{Object HalBench}} &\multicolumn{3}{c|}{\textbf{ROPE}}&\multicolumn{2}{c|}{\textbf{MMHal-Bench}}& \multicolumn{4}{c|}{\textbf{AMBER}}&\multicolumn{1}{c}{\textbf{{\footnotesize LLaVA-Bench}}}&\multicolumn{1}{c}{\textbf{MMBench}}\\
& &&$\text{Rsp.}^{\downarrow}$ & $\text{Men.}^{\downarrow}$ &$\text{Wild}^{\uparrow}$ & $\text{Hom}^{\uparrow}$ & $\text{Het}^{\uparrow}$& $\text{Score}^{\uparrow}$ & $\text{Hal}^{\downarrow}$ & $\text{Acc.}^{\uparrow}$& $\text{F1}^{\uparrow}$& $\text{CHAIR}^{\downarrow}$&$\text{Hal}^{\downarrow}$&$\text{Overall}^{\uparrow}$ &$\text{Overall}^{\uparrow}$ \\
\midrule
VCD \cite{VCD}&7B &  \ding{55} &48.8 &24.3&-&-&- & 2.12& 54.2&78.1&74.9&-&-&65.8&- \\
Yi-VL \cite{young2024yi}&6B &  \ding{55} &- &-&2.95&5.65&1.99 &- &- &-&-&-&-&51.9&65.6 \\
MiniGPT-4 \cite{zhu2023minigpt4} &7B &  \ding{55} &28.9 &20.4&-&-& -& 1.41& 76.2&63.6&64.7&13.6&65.3&45.1&32.7 \\
QWEN-VL \cite{bai2023qwen} &10B &  \ding{55} &40.4 &20.7 &2.73&6.60&1.03& 2.76& 38.5&81.9&86.4&5.5&23.6&71.9 &59.5\\
LLaVA-NeXT~\cite{llavanext}   & 34B  & \ding{55} & 12.6 & 6.4 &-&-&-& \textbf{3.31} & 34.4 & 81.4&85.4&-&-&77.7 &- \\
 Less-is-more \cite{yue2024less}&7B&\ding{55}& 40.3&17.8&18.1&
38.5&5.45& 2.33 &50.0&72.4& 75.8&5.7&24.4&-&-\\
 OPERA \cite{huang2024opera}&7B&\ding{55}& 45.1& 22.3&-&-&-&2.15 &54.2&-&-&-&-&-&-\\
\midrule
HA-DPO~\cite{ha_dpo} & 7B & Rule & 39.9 & 19.9 &17.0 &38.5&5.81& 1.98 & 60.4  &75.2&79.9&6.2 &28.2& 67.2 &- \\
BPO~\cite{pi2025strengthening}  & 7B & Rule & 30.6 & 16.4& 23.2&
47.7&8.37&  2.01 & 54.2  &  68.2&75.6&2.3&12.6&71.6 &- \\
 AMP-MEG \cite{zhang2024automated}&13B&Rule& 31.7 &20.6&-&-&-& 3.08& 36.5& 79.5& 84.6&10.0&45.7&-&-\\
\midrule
LLaVA-RLHF~\cite{llavarlhf} & 13B & Human & 38.1 & 18.9 &3.37&8.64&2.15 & 2.02 & 62.5 & 79.7&83.9&-&-&61.5&- \\
RLHF-V~\cite{2023rlhf-v}   & 13B & Human & 12.2 & 7.5 &-&-& -& 2.45 & 51.0 & 72.6&75.0&-&-&51.4&-\\
\midrule
Silkie~\cite{2023vlfeedback} & 10B & GPT-4V & 27.1 & 13.4&-&-&- & 3.19 & 32.3 & 82.2&\underline{87.6}&-&-&73.2 &-\\
RLAIF-V\cite{yu2024rlaif} & 7B & {\small{\footnotesize LLaVA-NeXT}} & 10.5 & 5.2 &18.9&48.8&5.81& 2.95 & 32.3 & 76.8&84.5&\underline{3.0}&\underline{16.4}&64.9&62.8 \\
\midrule
LLaVA-v1.5~\cite{llava15} & 7B & \ding{55} & 53.6 & 25.2 &13.9&31.8 &3.98& 2.36 & 51.0 & 72.0&74.7&7.8&36.4&59.7&62.5 \\
\rowcolor{Gray!20}
+ Ours & 7B & Self & \textbf{4.74} & \textbf{2.59} &23.5&48.7&7.40& 2.71 & 44.8 &80.4 &85.2&\textbf{2.0}&\textbf{10.3}&61.4&64.2 \\
LLaVA-NeXT~\cite{llavanext} & 7B & \ding{55} &17.3&9.50&31.8&71.7&11.5& 2.79&43.8&76.5&80.3&7.9&49.5&68.1&63.1\\
\rowcolor{Gray!20}
+ Ours & 7B & Self & \underline{4.87}&\underline{3.42}&33.0&\underline{74.2}&12.0&2.97&\textbf{36.5}&\underline{81.7}&85.7&5.5&32.4 &69.0&68.3\\
LLaVA-LLaMA3~\cite{grattafiori2024llama}& 8B & \ding{55} &12.1&7.43&36.1&74.1&14.1&2.68&49.0& 78.2&82.0&5.2&30.3&69.2&72.1\\
\rowcolor{Gray!20}
+ Ours & 8B & Self &7.72& 4.57& 37.6&\textbf{75.0}&15.5&2.86&43.8&79.8&83.2&4.0&23.5&70.7&72.9\\

Qwen2.5 VL\cite{bai2025qwen2}&7B&\ding{55}&15.1&8.20&\underline{48.9}&72.5&\underline{33.7}&\underline{3.15}&39.6&80.6&86.5&6.3&26.6&77.5&82.4\\
\rowcolor{Gray!20}
+ Ours&7B&self& 11.5&6.05&\textbf{49.5}&72.6&\textbf{34.9}&\textbf{3.30}&\underline{36.5}&\textbf{84.7}&\textbf{89.2}&5.0&25.2&79.0&82.5\\
\midrule
\textcolor{gray}{GPT-4V}~\cite{GPT4V} & - & - & \textcolor{gray}{13.6} & \textcolor{gray}{7.3} &\textcolor{gray}{53.8}&\textcolor{gray}{77.6}&\textcolor{gray}{40.8}& \textcolor{gray}{3.49} & \textcolor{gray}{28.1} & \textcolor{gray}{83.4} &\textcolor{gray}{ 87.4 }&\textcolor{gray}{4.6}&\textcolor{gray}{30.7}& \textcolor{gray}{93.1}&\textcolor{gray}{74.3} \\

\bottomrule
\end{tabular}}
\end{table*}

\subsection{Main Results}
We evaluate model performance in terms of both object hallucination and semantic hallucination. For object hallucination (Section~\ref{sec:obj_hal}), we assess the accuracy of object existence claims in open-ended captioning responses and binary (\emph{yes-or-no}) question-answering responses. 
For semantic hallucination (Section~\ref{sec:semantic_hal}), we analyze errors in attributes, relations, and other semantic details in the QA outputs. 
In addition, we incorporate instruction-following and QA benchmarks to validate the overall robustness of our method as in Section~\ref{sec:instruction_following}.

\subsubsection{Object Hallucination in VLMs}
\label{sec:obj_hal}
We compare our method against state-of-the-art approaches in Table \ref{tab:main}, evaluating both generative (captioning) and discriminative (QA) object hallucination metrics.

\noindent\textbf{Evaluation on captioning.}
We evaluate generative object hallucination using Object HalBench and AMBER-Generative.
For Object HalBench, our method with LLaVA-v1.5-7B achieves the best performance across all models, obtaining 4.74 in Rsp. and 2.59 in Men., outperforming larger models such as LLaVA-RLHF and RLHF-V as well as methods leveraging feedback from more advanced models, including RLAIF-V and Silkie.
For AMBER-Generative, as shown in Table~\ref{tab:amber}, our LLaVA-v1.5-7B variant achieves a CHAIR score of 2.0 and a hallucination rate of 10.3\%, while maintaining strong coverage at 50.2. Compared to our baseline (CHAIR: 6.5, hallucination rate: 29.4\%, coverage: 51.4), our approach effectively reduces hallucinations while retaining high factual coverage. Additionally, our model surpasses RLAIF-V, which relies on feedback from the significantly larger LLaVA-NeXT-34B, further demonstrating the strength of self-reflection preference optimization in mitigating hallucinations.


\noindent\textbf{Evaluation on QA.}
We evaluate object hallucination in question-answering using the Existence task from AMBER-Discriminative and the ROPE benchmark~\cite{chen2025multi}, which assesses hallucination in multi-object scenarios.
As shown in Table~\ref{tab:amber}, our approach consistently improves over the baseline for both LLaVA-v1.5-7B (F1 score: 83.3 $\rightarrow$ 94.1) and LLaVA-NeXT-7B (F1 score: 87.6 $\rightarrow$ 93.7).
For ROPE, we evaluate on the unseen split and report accuracy across different object distributions: \textit{Wild} (objects from mixed categories), \textit{Hom} (objects from the same category), and \textit{Het} (objects from distinct and unrelated categories, the most challenging setting).
Our method outperforms the baseline across all settings, with notable improvements in \textit{Wild} and \textit{Hom} cases. For instance, LLaVA-NeXT-7B improves from 31.8 to 33.0 on \textit{Wild}, while LLaVA-v1.5-7B improves from 13.9 to 23.5 on \textit{Wild} and from 3.98 to 7.40 on \textit{Het}, outperforming other 7B models using the same LLaVA-v1.5-7B backbone.
These results highlight the robustness of our method in mitigating hallucinations, even in challenging multi-object reasoning scenarios.

\begin{table*}[htp!]
\caption{Detailed evaluation results on AMBER~\cite{wang2024amber}. The generative evaluation assesses object hallucination in captions, while the Existence, Attribute, and Relation tasks measure incorrect object presence, misattributed properties (\eg, color, shape, or quantity), and inaccurate spatial or relational descriptions, based on short binary question responses. Since the questions of Existence tasks are all negative responses, recall and accuracy are equivalent; thus, we omit accuracy here. All models compared against are based on LLaVA-v1.5-7B to ensure a fair comparison.}
    \centering
    \small
\setlength{\tabcolsep}{1.5mm}
\begin{tabular}{c|cccc|ccc|cccc|cccc}
\toprule
 \multirow{2}{*}{\textbf{Model}}   &\multicolumn{4}{c|}{\textbf{Generative}}&\multicolumn{3}{c|}{\textbf{Existence}}&\multicolumn{4}{c|}{\textbf{Attribute}}&\multicolumn{4}{c}{\textbf{Relation}}\\
& $\text{CHAIR}^{\downarrow}$&$\text{Cover}^{\uparrow}$&$\text{Hal}^{\downarrow}$&$\text{Cog}^{\downarrow}$&$\text{P.}$& \text{R.}&$\text{F1}^{\uparrow}$ & $\text{Acc.}^{\uparrow}$ &$\text{P.}$&\text{R.}&$\text{F1}^{\uparrow}$ &$\text{Acc.}^{\uparrow}$ &$\text{P.}$&\text{R.}&$\text{F1}^{\uparrow}$  \\
\midrule
AMP-MEG 7B\cite{zhang2024automated}&12.2&53.2&56.0&5.2&100&56.6&72.2&73.3&87.2&54.7&67.2&67.1&58.4&71.4&64.2 \\
BPO~\cite{pi2025strengthening}&2.3&51.7&12.6&0.7&100&72.6&84.1&69.6&68.5&72.7&70.5&48.6&44.5&96.5&60.9\\
Less-is-more \cite{yue2024less}&5.7&50.7&24.4&2.3&100& 81.7&89.9&76.4&81.0&68.9&74.5&61.5&51.9&97.5&67.7\\
LLaVA-RLHF 7B \cite{llavarlhf}&9.1&51.8&44.5&4.7&100&62.6&77.0&66.3&64.0&74.6&68.9&51.5&45.5&87.4&59.8\\ 
RLAIF-V \cite{yu2024rlaif}&3.0&51.4&16.4&1.0&100&90.4&94.9&68.4&69.9&64.4&67.0&51.4&43.7&60.1&50.6\\ \midrule
LLaVA-v1.5~\cite{llava15}&7.8&51.0&36.4&4.2&100&71.5&83.3&72.0&88.0&51.0&64.6&73.9&72.0&60.2&65.6\\
+ Ours&\textbf{2.0}&50.2&\textbf{10.3}&\textbf{0.5}&100&91.7&\textbf{95.6}&76.5&76.5&76.5&76.5&64.5&54.5&86.8&67.0\\
LLaVA-NeXT~\cite{llavanext}&7.9&\textbf{63.1}&49.5&4.7&100&79.1&88.3&76.5&83.7&65.9&73.7&68.6&63.4&57.0&60.0\\
+ Ours&5.5&59.7&32.4&1.7&100&88.2&93.7&\textbf{78.4}&80.6&74.8&\textbf{77.6}&\textbf{77.6}&67.5&88.5&\textbf{76.6}\\
\bottomrule
\end{tabular}

\vspace{-2mm}
\label{tab:amber}
\end{table*}

\subsubsection{Semantic Hallucination in VLMs}
\label{sec:semantic_hal}
We evaluate semantic hallucination using MMHal-Bench and AMBER-Discriminative, which assess various semantic aspects including object attributes, comparisons, counting, relations, and environmental context.
On MMHal-Bench (Table~\ref{tab:main}), our variants based on both LLaVA-v1.5 and LLaVA-NeXT show improvements over the baseline, with scores increasing from 2.36 to 2.71 and from 2.79 to 2.97, respectively, while reducing the hallucination rate from 51.0\% to 44.8\% (LLaVA-v1.5) and from 43.8\% to 36.5\% (LLaVA-NeXT). 
Notably, our LLaVA-v1.5 variant outperforms rule-based methods such as HA-DPO~\cite{ha_dpo} and BPO~\cite{pi2025strengthening} , although it is slightly outperformed by RLAIF-V ~\cite{yu2024rlaif} (2.71 vs. 2.95), whose feedback is derived from a larger LLaVA-NeXT-34B model.
Among 7B models, our LLaVA-NeXT-7B variant achieves the best score of 2.97
In the Attribute and Relation tasks of AMBER-Discriminative, our approach consistently improves the F1 score compared to the baseline, \eg, Attribute F1 increases from 64.6 to 77.2 for LLaVA-v1.5 and from 56.4 to 77.6 for LLaVA-NeXT, except for the relation task on LLaVA-v1.5, where performance remains similar (65.6 to 64.2). 


\subsubsection{Instruction-following Evaluation}
\label{sec:instruction_following}
We evaluate the instruction-following and general question-answering capabilities of our models to validate their robustness using LLaVA-Bench~\cite{llava15} and MMBench~\cite{liu2024mmbench}. 
As shown in Table~\ref{tab:main}, our variants improve instruction-following performance: LLaVA-v1.5-7B increases from 59.7 to 61.4, and LLaVA-NeXT-7B from 68.1 to 69.0. Furthermore, our method yields notable gains on MMBench, with performance improvements from 62.5 to 64.2 for LLaVA-v1.5-7B and from 63.1 to 68.3 for LLaVA-NeXT-7B. These results validate the effectiveness of our approach for ensuring strong instruction-following capability. 

\subsection{Ablations}
\label{sec:ablations}

In this section, we conduct ablation studies to assess the impact of key design choices in our framework, including different tasks for data generation, two-stage prompting, and data ranking measurement. We also examine how response length affects hallucination rates. 


\subsubsection{\textcolor{black}{Data sources}}

\noindent\textbf{Captioning \extmf{data}.} As shown in Table~\ref{tab:ab_overall}, comparing (i) and (ii), we observe that incorporating the captioning task leads to significant improvements across both generative and discriminative evaluations. Notably, the hallucination rate in generated captions drops from 29.4\% to 12.5\%, while CHAIR reduces from 6.5 to 2.5. Additionally, the discriminative performance benefits as well, with the F1 score improving from 77.7 to 83.3, highlighting the effectiveness of leveraging object existence consistency for improving factual grounding.

\begin{table}[t!]
\caption{Ablation on training data. Each row indicates whether the model is trained with captioning and/or QA data. We evaluate on Acc. and F1 from AMBER-Discriminative; CHAIR and Hal from AMBER-Generative.}
\centering
\small
\setlength{\tabcolsep}{2.2mm}
\begin{tabular}{c|cc|cccc}
    \toprule
&\multirow{2}{*}{Captioning}&\multirow{2}{*}{QA} &\multicolumn{4}{c}{AMBER}\\ 
&&&$\text{Acc.}^{\uparrow}$&$\text{F1}^{\uparrow}$&$\text{CHAIR}^{\downarrow}$&$\text{Hal}^{\downarrow}$ \\
\hline
i. &\ding{55} & \ding{55} & 73.5&77.7&6.5&29.4\\
ii. &\ding{51} & \ding{55} & 79.0&83.3&2.5&12.5\\
iii.&\ding{55} & \ding{51} &75.1&79.9&7.3&33.6 \\
\hdashline
iv.&\ding{51} & \ding{51} & \textbf{80.4}&\textbf{85.2}&\textbf{2.0}&\textbf{10.3}\\
    \bottomrule
\end{tabular}

\label{tab:ab_overall}  
\end{table}



\begin{figure*}[t!] 
    \centering
    \includegraphics[width=0.85\textwidth]{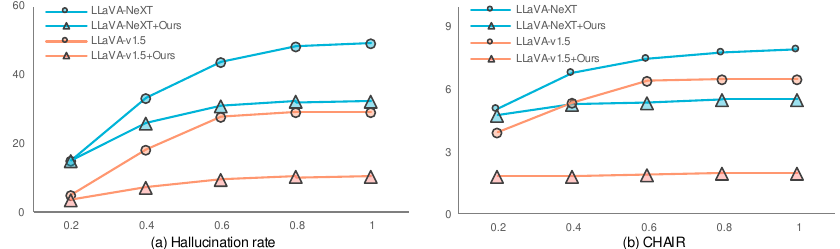}
    \vspace{-2mm}
    \caption{\textbf{Hallucination trends with respect to relative response length.}
    (a) and (b) show the performance of the response-level hallucination rate and the mention-level CHAIR metric, respectively, on the AMBER generative task.
    Different backbone models are distinguished by color, while different marker shapes denote the base models and our variants.
    }
    \label{fig:hal_trends}
    \vspace{-2mm}
\end{figure*}

\noindent\textbf{QA \extmf{data}.} 
As shown in Table~\ref{tab:ab_overall}, comparing (i) and (iii), introducing the QA data improves discriminative hallucination detection, boosting the F1 score from 77.7 to 83.3 and Acc. from 73.5 to 79.0. However, it also increases hallucination in the model’s generative captions, leading to a degradation in Hal from 12.5\% to 33.6\% and CHAIR from 2.5 to 7.3. We suspect that this performance drop is potentially due to the increased complexity of broad semantic inconsistencies.

Interestingly, when combining both captioning and QA data (comparing (ii) and (iv)), the QA data contributes positively to the captioning task, further reducing the hallucination rate from 12.5\% to 10.3\% and CHAIR from 2.5 to 2.0, while also improving the F1 score from 83.3 to 85.3. This demonstrates the complementary nature of captioning and  QA data, where the strengths of each task help mitigate their respective weaknesses, leading to a more robust hallucination detection and reduction strategy.

\begin{table}[tb!]
\caption{Ablation within the captioning data only. Columns indicate whether the data ranking uses object claims, semantic aspects, or both. We evaluate on Acc. and F1 from AMBER-Discriminative; CHAIR and Hal from AMBER-Generative.}
\centering
\small
\setlength{\tabcolsep}{2.1mm}
\begin{tabular}{c|cc|cccc}
    \toprule
&\multirow{2}{*}{\shortstack{Object\\claims}}&\multirow{2}{*}{\shortstack{Semantic\\aspects}} &\multicolumn{4}{c}{AMBER}\\ 
&&&$\text{Acc.}^{\uparrow}$&$\text{F1}^{\uparrow}$&$\text{CHAIR}^{\downarrow}$&$\text{Hal}^{\downarrow}$ \\
\hline
i. &\ding{55} & \ding{55} & 73.5&77.7&6.5&29.4\\
\hdashline
ii. &\ding{51} & \ding{55} & \textbf{79.0}&\textbf{83.3}&\textbf{2.5}&\textbf{12.5}\\
iii. &\ding{51} & \ding{51} & 78.9&83.2&3.0&15.2\\
    \bottomrule
\end{tabular}

\label{tab:ab_cap}  
\end{table}

\begin{table}[t!]
\vspace{-1mm}
\caption{Impact of ranking by coverage during preference data curation on AMBER~\cite{wang2024amber}. Using the coverage as a ranking index improves Gen;Coverage but leads to higher Gen;Hal and lower Disc;F1, revealing a trade-off between completeness and hallucination suppression.}
\small
    \centering
    \setlength{\tabcolsep}{1.8mm}{
    \begin{tabular}{l c c c}
        \toprule
        Method & Gen;Coverage$\uparrow$ & Gen;Hal$\downarrow$ & Disc;F1$\uparrow$ \\
        \midrule
        LLaVA-v1.5~\cite{llava15} & 51.0 & 36.4 & 74.7 \\
        + \textbf{Ours}{(\scriptsize Inconsistency)} & 50.2 & \textbf{10.3} & \textbf{85.2} \\
        + \textbf{Ours}{(\scriptsize Coverage)} & \textbf{57.8} & 25.9 & 80.4 \\
        \bottomrule
    \end{tabular}
    }
\label{tab:coverage}
\end{table}

\subsubsection{\extmf{Atomic claim extraction}}

\noindent\textbf{Semantic aspects in captions.} 
We also explored extracting semantic atomic claims from captions to expand the data within the captioning dataset(without the QA data). \hhh{We follow same practice as in question-answering data to extract semantic claims.}
However, as shown in Table~\ref{tab:ab_cap}, comparing (ii) and (iii), incorporating semantic aspects into the inconsistency measurement does not enhance the model’s discriminative hallucination detection and instead leads to a degradation in generative performance, with the Hal. increasing from 12.5\% to 15.2\%. This is potentially due to the increased complexity of ranking responses based on broader semantic inconsistencies, and lower short response accuracy with semantic facts (Figure~\ref{fig:observation}). 

\noindent\textbf{Two-round v.s. Single-round prompting.} As illustrated in Section~\ref{sec:data_curation}, we adopt the two-round CoT prompting approach for the QA data to facilitate extraction of semantic aspects. To evaluate its effectiveness, we conduct an ablation experiment with single-round prompting (\ie, without question-relevant description first and directly answer the question).
It only achieves a significantly lower AMBER F1 score of 72.3, falling below the baseline score of 77.7. 
\hhh{We attribute this degradation to the short answer responses, which lack sufficient semantic facts for reliable self-consistency rating. 
}
\extmf{These results validate the effectiveness of our two‑round CoT response generation for QA data.}



\subsubsection{\extmf{Preference data curation}}

\noindent\textbf{Occurance v.s. Relative ratio for ranking.} 
We compare two ranking strategies for evaluating model consistency: absolute occurrences of inconsistencies and the normalized inconsistency ratio (\ie, occurrence normalized by the total number of extracted short questions).
The results of occurrence-based ranking with caption data only is reported in (ii) in Table~\ref{tab:ab_overall}. We further experiment with normalized ratio for ranking. 
While still outperforming the baseline, the relative ratio approach underperforms the occurrence-based ranking, yielding an F1 score of 80.3 (compared to 83.3) and a Hal. of 14.6\% (compared to 12.5).
We attribute this to the nature of hallucination itself: each occurrence of inconsistency is equally critical. The occurrence-based approach ensures that all factual errors contribute proportionally to ranking. In contrast, the normalized ratio can dilute the impact of severe hallucinations in longer responses.



\noindent\extmf{\textbf{Considering Claim Coverage.}} 
\extmf{Our preference ranking primarily considers the occurrence of inconsistent behavior in the model’s responses, without explicitly accounting for claim coverage. 
To examine the impact of incorporating coverage, we conducted an ablation on our LLaVA-v1.5 by enforcing an \textbf{additional} constraint during data curation: a candidate is only selected as the preferred response if it contains at least as many atomic claims as the rejected one.
This stricter criterion led to a 53\% reduction in training preference pairs. As shown in Table~\ref{tab:coverage}, enforcing claim coverage substantially improves object coverage (Gen;Coverage) (from 50.2 to 57.8) but also increases hallucination (Gen;Hal rises from 10.3 to 25.9) and a modest drop in overall discrimination F1 score (Disc;F1) from 85.2 to 80.4). These results on AMBER~\cite{wang2024amber} indicate a trade-off between factual completeness and hallucination removal. Given the sharper increase in hallucination, we adopt the simpler ranking strategy without the coverage constraint.}

\begin{figure*}[t!] 
    \centering
    \includegraphics[width=1.\textwidth]{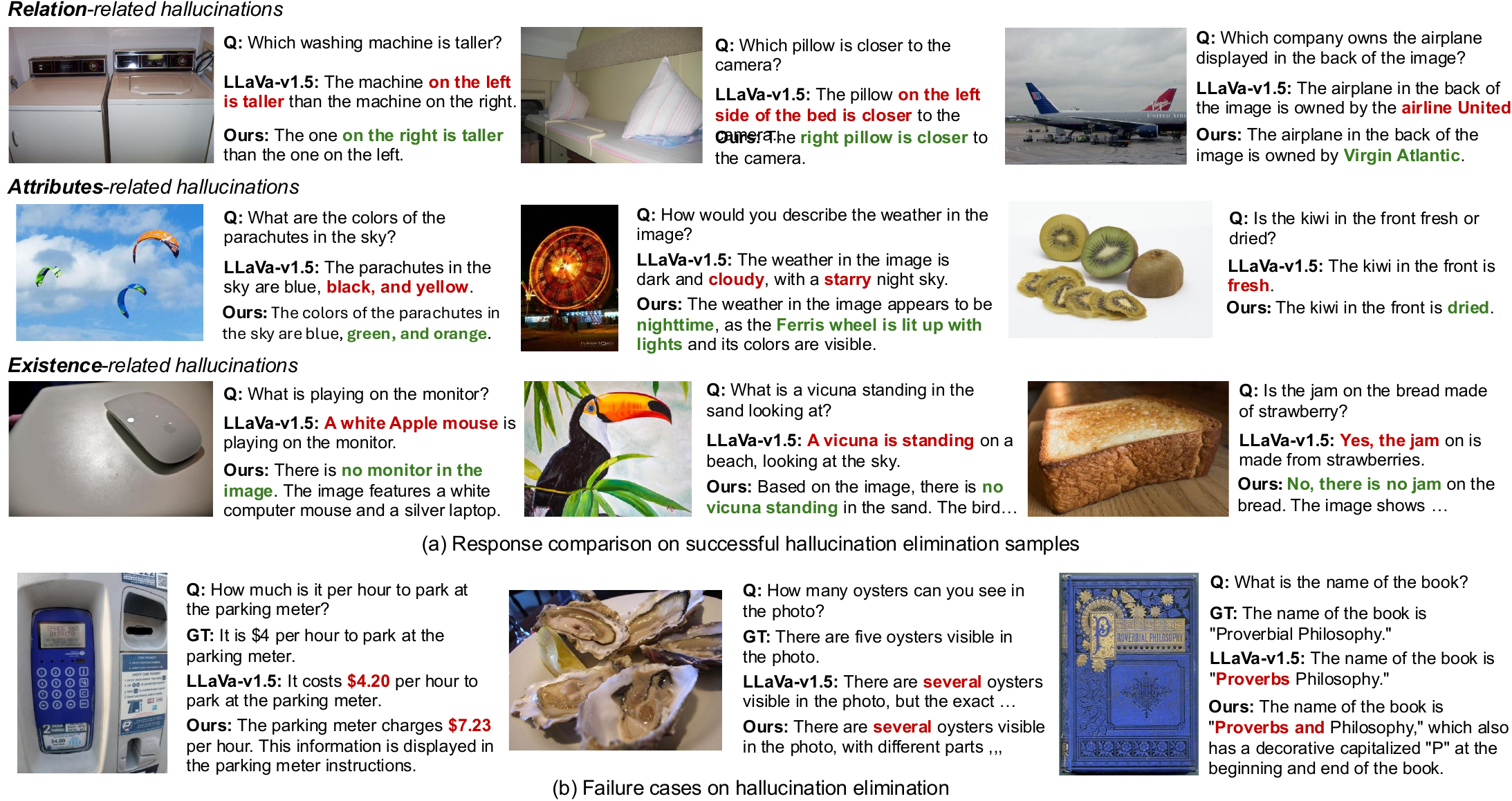}
    \vspace{-2mm}
    \caption{\extmf{\textbf{Qualitative comparison of hallucinations in LLaVA‑v1.5‑7B versus our method.}
    \textbf{(a) Successful eliminations:} across three categories—relation (\eg, correctly grounding which washing machine is taller, pillow position, airplane in the back), attribute (parachute colours, Ferris‑wheel weather, kiwi freshness), and existence (detecting absence of a monitor, vicuna, or jam even when hypothesized)—our model (green) removes the spurious content introduced by the baseline (red).
    \textbf{(b) Remaining failures:} both models still struggle with OCR/text reading (parking‑meter fee), exact enumeration (oyster count), and fine‑grained text recognition (book title). Please zoom in for high‑resolution details.
    }}
    \label{fig:vis_samples}
\end{figure*}

\subsubsection{\extmf{Hallucination rate over generated token length.}}
 \hhh{In order to measure the relationship between hallucination and the generated response length, we experiment with responses in different lengths as shown in Figure~\ref{fig:hal_trends}. }
For a fair comparison, the response lengths with the same backbone are normalized using the same factor.
Longer outputs generally exhibit increased hallucinations.
With our self-reflection preference optimization, the growth in hallucination rate is significantly mitigated.
For instance, when the response length is increased to 80\% of the maximum, our variant based on LLaVA-NeXT exhibits a lower hallucination rate than the base model at 40\% length.
Similarly, the CHAIR metric shows significant improvement in our variants compared to the base models, particularly with the response length increases beyond 20\%. Importantly, our variants still achieve comparable coverage with these gains in reducing hallucinations, as shown in Table~\ref{tab:amber}. 

\subsection{\extmf{Qualitative Visualization}}
We provide qualitative visualization in Figure~\ref{fig:vis_samples} to illustrate how effectively our method mitigates hallucinations generated, \eg, by LLaVA-v1.5-7B, across three major hallucination categories: \textit{relation}, \textit{attribute}, and \textit{existence}.

Figure~\ref{fig:vis_samples}(a) demonstrates successful hallucination elimination examples. In the \textit{relation} category, our model correctly identifies and grounds spatial or positional relationships between objects. For instance, when queried about which washing machine is taller, our model accurately identifies the right machine, contrary to the baseline which incorrectly selects the left one. Within the \textit{attribute} category, our method effectively corrects mistakes related to intrinsic properties or attributes of objects. Examples include accurate recognition of parachute colors, correctly identifying the weather conditions around a Ferris wheel (nighttime illumination rather than an incorrectly inferred starry sky). For the \textit{existence} category, our approach robustly eliminates hallucinations arising from the incorrectly hypothesized presence of objects. For example, the baseline incorrectly describes a mouse as playing on a non-existent monitor, identifies an absent vicuna standing in the sand, and falsely claims the presence of strawberry jam on bread. Our model successfully addresses these existence-related hallucinations by explicitly acknowledging and rejecting these non-existent elements, underscoring its improved behavior consistency and visual grounding capabilities.

Nevertheless, our method still exhibits limitations, as shown in Figure~\ref{fig:vis_samples}(b). Both our approach and the baseline fail at tasks involving detailed OCR/text reading, such as accurately identifying the exact parking meter fee. Additionally, enumeration tasks, like counting oysters precisely, remain challenging due to intrinsic difficulties in model enumeration ability without external training signals. Moreover, fine-grained text recognition tasks, exemplified by extracting the exact title of a book (``Proverbial Philosophy''), also highlight persistent shortcomings.
We attribute these failure cases to the intrinsic limitation of our method, as our approach does not incorporate any external supervision or additional specialized training data. 

\section{Conclusions}
In this work, we introduced a self-reflection preference optimization framework designed to enhance factual grounding in vision-language models by reducing hallucinations. Our method uses an self-consistency signal, comparing detailed model responses with their corresponding short binary responses, to generate high-quality training data without human annotations or external model feedback. Extensive experiments across diverse benchmarks for both captioning and question-answering tasks demonstrate that our approach substantially reduces hallucinations and improves overall response quality as well as instruction-following capabilities. \extmf{Furthermore, our method generalizes effectively to multiple representative vision-language architectures, highlighting its robustness and practical applicability.}

\vspace{+2mm}
\noindent\textbf{Limitations \extmf{and Future Work}.} 
Although self-reflection is shown to be effective in reducing hallucinations for VLMs, there is still a significant portion of hallucinations that cannot be detected due to the limited capacity of the target model. \hhh{Meanwhile, our method’s pursuit of fine-grained factual accuracy can make the model’s responses more cautious and conservative, which may reduce the imagination and richness of the output.}
\extmf{In future work, we plan to integrate internal self-reflection mechanisms with external supervision and additional specialized training signals. Potential strategies include utilizing auxiliary training objectives, specialized data augmentation, external knowledge databases, explicit counting mechanisms, or pretrained OCR modules. Exploring adaptive ranking schemes that balance hallucination suppression with informative completeness could further enhance model performance. These efforts aim to ultimately broaden the applicability and reliability of vision-language models in real-world scenarios.}

\bibliographystyle{IEEEtran}
\bibliography{main}  

\clearpage
\setcounter{page}{1}
\onecolumn
\appendix
\section*{Structured Prompt for Claim Extraction}
We show the prompt used for relation- and attribute-based semantic claim extraction in Table~\ref{tab:prompt}. 
Without loss of generability, we employ GPT-4 with this structured prompt to return a JSON object response, which we then parse to atomic facts.
\begin{table*}[hbp]
 \caption{\hhh{Structural prompt used to extract different semantic atomic claims from responses. In extraction, we consider attributes, relationships, actions, etc., while avoiding duplication or ambiguity.}}
\begin{tcolorbox}[colback=lightgray!30, colframe=black, sharp corners=all, width=\textwidth, boxrule=0.5mm, title=Prompt for Atomic Claim Extraction]
\small 
\scriptsize
\# Extract key aspects of an image description and structure the output in JSON format, to assist in extracting our question. Because the description is too complex, I prefer to first extract the key aspects of the description to assist in extracting our questions. \\
\\
\# Requirement of key aspects:\\
-Content composition: The aspects I cared about were: knowledge and functionality (simple and short facts that are not specific to the image, \ie, general knowledege and statement), object quantity (Nouns such as people, animals, plants, and objects etc. that appear. \#I only care about objects with specific values, not objects with unclear values \#), relation between objects (Direct-contact object relations, interactive actions or positional relationships among objects), actions (non-interactive actions and behaviors, e.g., run, walk, take a shower, etc.), object attributes(Size, shape, color, material, etc.), image content reasoning (object quantity, object relation, object actions that are implicityly conveyed by the image description; it's okay to be empty), and others (such as texts in the image, clock reading, etc.).\\
-Content Requirements: Content should be distinct and non-overlapping as much as possible.\\
-Other Requirements: For each extracted item in a specific aspect, I require the original text from a description sentence in proper form and MUST be strictly consistent with the original sentence.\\
-Extract some simple questions (does not contain composite description) (yes-or-no answerable question) paired to key aspects item in JSON format.\\
\\
\# Requirement of questions:\\
-Content Requirements: The question needs to be a yes-or-no answerable question, paired to key aspects item in JSON format. Content should be distinct and non-overlapping as much as possible.\\
-Content Requirements: The objects in the question \#MUST be simple objects or simple relations, not complex ones. Multiple question should be decomposed.\\
-Format requirements: Question in "object quantity" iterm requires the use of $\langle$ multi $\rangle$ to indicate more than one of the same object with a specific number(e.g., two, three, four, etc.). \\
-Format Requirements: "object quantity": $\langle$ multi $\rangle$Are there specific number noun? "object relation": Are/Is noun relation noun? "object attributes": Are/Is noun attribute? "object action": Are/Is noun verb?
-Other Requirements: One aspect item can correspond to multiple questions, keep each question as concise as possible.
-Other Requirements: Question in "object\_quantity" iterm requires the use of <multi> to indicate more than one of the same object with a specific number(e.g., two, three, four, etc.). <multi> only appears in the "object\_quantity" iterm.\\
-Other Requirements: All vague (eg: \#some\#, \#several\#, \#a group of\#) or speculative (eg: \#possibly\#, \#appears\#, \#seems\#) statements MUST be eliminated.\\
\\
\# Step Tips:\\
-Ignore vague or speculative statements in the input.
-Check whether the requirements of aspects are met and remove unsatisfied items.\\
-Extract questions based on the extracted key aspects.
-Organize aspects and questions in the required format to obtain output.\\
-Check whether questions meet the requirements and remove unsatisfied items.\\
-Organize aspects and questions in the required format to obtain output.\\
\\
\# Examples\\
**Example Input**:"Yes, a dog on the right side of two women might want to pee on the yellow object, as it is a fire hydrant. Fire hydrants are typically found in public spaces, like parks or streets, and are used by firefighters to access water in case of emergencies. They are often painted in bright colors, such as yellow, to make them easily visible for both humans and animals. In this case, the yellow fire hydrant in the grassy area is an attractive spot for dogs to mark their territory or simply to relieve themselves."\\
**Example Analysis**:There are many complex aspect items in the article, so when extrating questions, they need to be decomposed. Thus, they are corresponding to multiple questions. "Several" is a vague statement, so ignore this sentence in the input. (Analysis does not require output.)\\
\\
**Example Output**\\
```json\\
\{"knowledge\_and\_functionality":[
    \{
      "text": "Fire hydrants are typically found in public spaces, like parks or streets, and are used by firefighters to access water in case of emergencies.",
      "question": [
          "Are fire hydrants found in public spaces?" "Are fire hydrants used by firefighters?"  "Are fire hydrants used to access water in case of emergencies?"]\},
    \{
      "text": "They are often painted in bright colors, such as yellow, to make them easily visible for both humans and animals",
      "question": [
          "Are fire hydrants often in bright colors?"
          "Are fire hydrants painted to be easily visible?"
      ]
    \},
  ],
\\
  "object\_quantity": [
    \{
      "text": "a dog on the right side of two women might want to pee on the yellow object",
      "question": [
          "Are there two women?"
          ]
    \},
  ],\\
  "object\_relation": [
    \{
      "text": "a dog on the right side of two women might want to pee on the yellow object",
      "question": [
          "Is there a dog on the right side of two women?"
          "Does the dog pee on the yellow object?"
      ]
\},
    \{
      "text": "the yellow fire hydrant in the grassy area",
      "question": [
          "Is the fire hydrant in the grassy area?",
          ]
    \}
  ],\\
  "object\_attributes": [
    \{
      "text": "the yellow fire hydrant in the grassy area ",
      "question": [
          "Is the fire hydrant yellow?"
            ]
    \},
  ],\\
  "reasoning": [
   \{
      "text": "an attractive spot for dogs to mark their territory or simply to relieve themselves",
      "question": [
          "Is the dog peeing on fire hydrant for marking territory?",
          "Is the dog peeing on fire hydrant for relieving themselves?"
          ]
    \},],\\
  "others": [
  ]
\}\\
'''\\
\\
\# Notes: \\
-Vague (eg: some, several etc.) or speculative (eg: possibly, appears, seems, might, Perhaps, Maybe, Could, likely, etc.) statements \#MUST be eliminated in both aspects and questions.\\
-The questions raised can be missing if you are not sure, but they must not be wrong. Make sure that all the questions raised meet the previous requirements, and is placed in the correct aspects catalogue.\\
-The objects in the question \#MUST be simple objects or simple relations, not complex ones. Multiple question should be decomposed.
\end{tcolorbox}
\label{tab:prompt}
\end{table*}

\end{document}